%% file: main.tex
\documentclass[10pt,twocolumn,letterpaper]{article}

\usepackage[pagenumbers]{cvpr} % To force page numbers, e.g. for an arXiv version

\input{preamble}

\definecolor{cvprblue}{rgb}{0.21,0.49,0.74}
\definecolor{LMUGreen}{HTML}{00883A} % LMU 
\definecolor{LMUGrey1}{HTML}{E6E6E7}
\definecolor{LMUGrey2}{HTML}{C0C1C3}
\definecolor{LMUGrey3}{HTML}{626468}
\definecolor{LightPurple}{HTML}{969FFF}
\definecolor{Orange}{HTML}{FF8B00}
\definecolor{Pink}{HTML}{FF0090}

\usepackage{adjustbox}
\usepackage{siunitx}
\usepackage{url}
\usepackage{pgfgantt}
\usepackage{pgfplots}
\pgfplotsset{compat=1.17} 
\usepackage{multirow}
\usetikzlibrary{fit, arrows.meta, positioning, backgrounds}
\usepackage[pagebackref,breaklinks,colorlinks,citecolor=cvprblue]{hyperref}

\title{\textsc{GiMeFive}: Towards Interpretable Facial Emotion Classification}
\author{Jiawen Wang$^\dagger$ $^\star$\\ % $^\heartsuit$ 
$^\dagger$Institute of Informatics $\:$ $^\star$CIS\\
University of Munich, Germany\\
{\tt\small Jiawen.Wang@campus.lmu.de}
\and
Leah Kawka$^\dagger$\\
$^\dagger$Institute of Informatics\\
University of Munich, Germany\\
{\tt\small Leah.Kawka@campus.lmu.de}
}

\begin{document}
\maketitle
\input{sec/0_abstract}    
\input{sec/body}

{
    \small
    \bibliographystyle{ieeenat_fullname}
    \bibliography{main}
}

\input{sec/appendix}

\end{document}

%% file: preamble.tex
\usepackage[dvipsnames]{xcolor}

%% file: sec/0_abstract.tex
\begin{abstract}
    Deep convolutional neural networks have been shown to successfully recognize facial emotions for the past years in the realm of computer vision. 
    However, the existing detection approaches are not always reliable or explainable, 
    we here propose our model GiMeFive with interpretations, 
    i.e., via layer activations and gradient-weighted class activation mapping. 
    We compare against the state-of-the-art methods to classify the six facial emotions. 
    Empirical results show that our model outperforms the previous methods in terms of accuracy on two Facial Emotion Recognition 
    (FER) benchmarks and our aggregated FER GiMeFive. 
    Furthermore, we explain our work in real-world image and video examples, as well as real-time live camera streams. 
    Our code and supplementary material are available at \url{https://github.com/werywjw/SEP-CVDL}. 
\end{abstract}

%% file: sec/body.tex
\section{Introduction}
\label{sec:intro}

In recent years, facial recognition technology has matured significantly. 
As an important branch of facial recognition, 
\textit{Facial emotion recognition} (FER)~\cite{Ko18,JainSS19,YinTLS019,Malik0R21} has received growing attention from researchers. 
This progress is largely driven by the effective utilization of \textit{convolutional neural networks} (CNNs). 
Originally introduced for image classification, 
CNNs have proven their ability to capture spatial hierarchies, 
making them ideal for analyzing facial features and expressions. 
Since the pioneering work by \citet{Lecun89Backpropagation}, 
CNNs have significantly improved, fueled by larger datasets, 
powerful GPU implementations, 
and innovative regularization strategies. 

In the era of deep learning and \textit{artificial intelligence} (AI), 
one major focus in the research community has been on designing network architectures 
and objective functions towards discriminative feature learning~\cite{HeZRS16, liu2017learning, wen2016discriminative}. 
Simultaneously, 
there is a strong demand from researchers and general audiences to interpret its successes and failures~\cite{GoodfellowSS14}, 
and to understand, improve, and trust the generated decisions by these models. 
This has led to the development of 

\begin{figure}[ht]
  \centering
  \resizebox{.47\textwidth}{!}{
  \begin{tikzpicture}
    \begin{axis}[
        ybar, 
        legend style={at={(1.35,0.795)},
          anchor=east}, % ,legend columns=-1
        ylabel={Test Accuracies (\%)}, 
        symbolic x coords={VGG,ResNet18,ResNet34,Baseline,GiMeFive},
        x tick label style={rotate=30,anchor=east},
        every axis plot/.append style={
          bar shift=0pt
        },
        bar width=18pt, 
        ymin=78, 
        ymax=88,
        grid=major, 
        % ymajorgrids=true,
        % xmajorgrids=true,
    ]
    \addplot[fill=Pink!30] coordinates {(VGG,83.3)};
    \addlegendentry{VGG}
    \addplot[fill=LightPurple!30] coordinates {(ResNet18,81.3)};
    \addlegendentry{ResNet18}
    \addplot[fill=cvprblue!30] coordinates {(ResNet34,83.7)}; 
    \addlegendentry{ResNet34}
    \addplot[fill=Orange!30] coordinates {(Baseline,80.6)};
    \addlegendentry{Baseline}
    \addplot[fill=LMUGreen!30] coordinates {(GiMeFive,86.5)};
    \addlegendentry{\textsc{GiMeFive}}
    \end{axis}
  \end{tikzpicture}
  }
  \caption{Test accuracies (\%) of our \textsc{GiMeFive} 
  compared to other state-of-the-art models on the RAF-DB dataset (see \cref{tab:model} for full results).} 
  \label{fig:acc}
\end{figure}
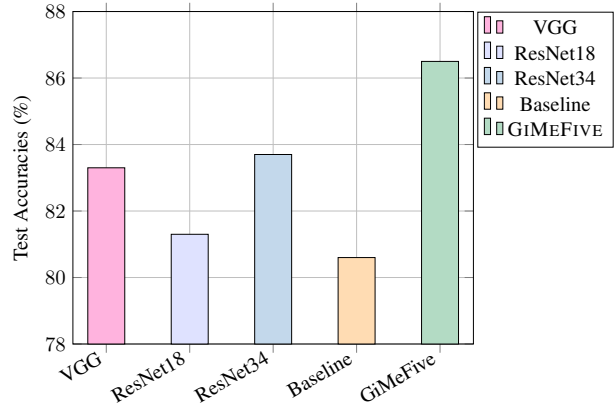

various tools for visualizing CNNs and dissecting their prediction paths to identify the important visual cues they rely on~\cite{olah2018building}. 

Facial expressions, a cornerstone of non-verbal communication, 
hold significant weight in human interactions. 
Leveraging \textit{explainable AI} (XAI)~\cite{olah2018building,YinTLS019,Malik0R21}, 
we shed light on the decision-making process behind our emotion classifier, 
enhancing its transparency. 
In this report, we present our best model named \textsc{GiMeFive}, 
besides several deep CNNs to detect and interpret six basic universally recognized and expressed human facial emotions, 
namely happiness, surprise, sadness, anger, disgust, and fear. 
The goal of our research is to build a robust and interpretable model for facial emotion recognition tasks. 

\begin{figure*}[ht]
  \centering
   \includegraphics[width=\linewidth]{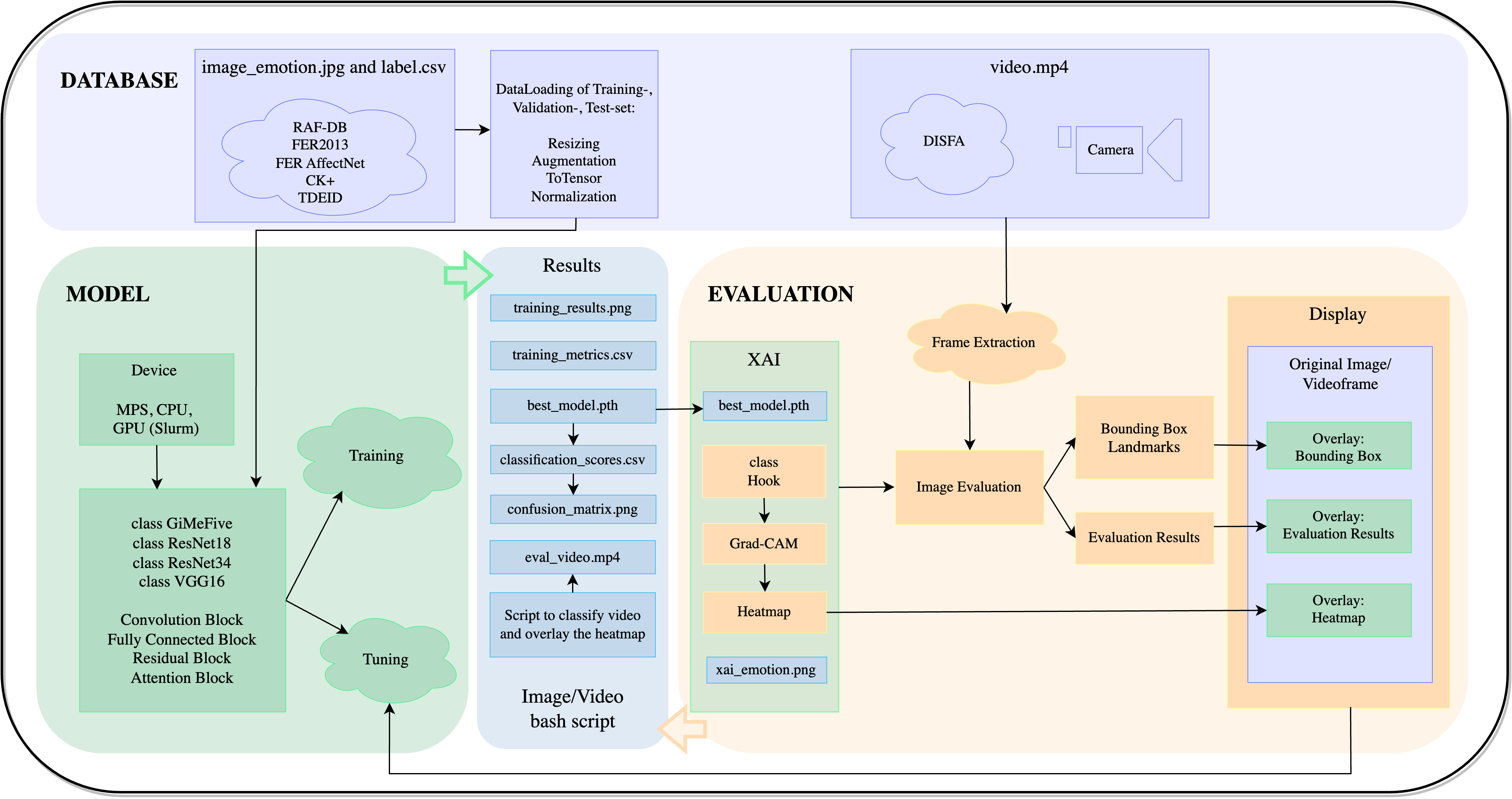}
   \caption{Overview of the experimental pipeline.} 
   \label{fig:pipeline}
\end{figure*}

Our main contributions can be summarized as follows. 
\begin{itemize}
  \item We collect, preprocess, and evaluate the training and testing data 
  (both images and videos) from various public databases thoroughly. 
  \item We implement all classification models from scratch and optimize them with several techniques in a systematic manner. 
  Meanwhile, we illustrate the classification scores with the confusion matrix heat map. 
  \item We give several video demos to show the real-world performance of our best model \textsc{GiMeFive}. 
  \item We provide qualitative benefits such as interpretability to explain our model with gradient-weighted class activation mapping and face landmarks.
\end{itemize}

\paragraph{Paper Outline.}
The structure of the rest of the report is arranged as follows. 
\Cref{sec:related} contains the related work of our research. 
In \Cref{sec:setup}, 
we address the datasets we collected and the model architecture we implemented. 
An overview of the experimental pipeline of our project is outlined in \Cref{fig:pipeline}. 
The evaluation results of our models are given in \Cref{sec:evaluation} with interpretability. 
The test accuracy on the RAF-DB dataset is shown in \Cref{fig:acc}. 
\Cref{sec:optim} describes the optimization strategies such as data augmentation and hyperparameter tuning. 
We provide the conclusion and discussion in \Cref{sec:conclusion}. 

\section{Related Work}
\label{sec:related}

% Facial Expression Recognition with Adaptive Frame Rate based on Multiple Testing Correction
% https://github.com/kdhht2334/awesome-SOTA-FER

\paragraph{Facial Emotion Recognition.}
% Previous research by \citet{SamadianiHCLCXH19} and \citet{mehrabian2017communication} confirms this, 
% highlighting their substantial contribution to communication. 
Studies from \citet{ekman1971constants} and \citet{LiD22a} categorize facial expressions into distinct states 
such as happiness, sadness, or anger, 
demonstrating their cross-cultural recognition. 
While emotions can intertwine in specific situations, 
primary emotions remain broadly identifiable due to their innate nature.
Facial emotion recognition capitalizes on this understanding by classifying the emotional states of individuals based solely on their facial expressions. 
This field has witnessed a shift from hand-crafted features to sophisticated deep learning architectures, 
achieving remarkable accuracy on benchmark datasets~\cite{FardM22,LiGL22}. 
% In any case, ethical considerations are paramount. 
Given the widespread use of video streams and FER applications, 
real-time operation is crucial. 
However, 
this must be balanced with fairness, robustness, and scalability, 
aspects often overlooked in recent research~\cite{Savchenko22}. 
\citet{WangQKGNHR20} proposed a benchmark for bias mitigation, 
while \citet{PhamVT20} presented a novel masking idea to boost the performance on the FER2013~\cite{BarsoumZCZ16} and \textit{Vietnam Emotion} (VEMO) dataset. 
% Building upon recent, comprehensive reviews of FER models and datasets~\cite{LiGL22,CanalMMSSPS22,sajjad2023comprehensive}. 
% our research delves into a range of CNNs with varying complexity. 
% We utilize datasets with extensive labeling encompassing subpopulations, head poses, occlusion types, and class imbalances. 
% This approach allows us to explore the intricate relationship between facial expressions and emotions while ensuring responsible development and application of FER technologies.
% By combining the power of facial expressions with advanced machine learning, 
% FER has the potential to revolutionize various fields. 
% However, ethical considerations must be woven into the fabric of this technology from the outset to guarantee its responsible and inclusive use in shaping the future of human-computer interaction.
% while Xu et al. (2020) focused on sensitive attributes like gender, age, and race.
% cite did not found

% Peeking Inside the Black-Box: A Survey on Explainable Artificial Intelligence (XAI)
% Explainable Artificial Intelligence (XAI): Concepts, taxonomies, opportunities and challenges toward responsible AI
\paragraph{Explainable AI.}

Understanding the visual world through machines has been a driving force in computer vision for decades~\cite{SudderthTFW05,parikh2011human,SinghGE12,JunejaVJZ13,MahendranV16}. 
Early attempts at peering into the ``black box'' of recognition models included directly visualizing filter responses~\cite{ZeilerF14}, 
reconstructing inputs from network layers~\cite{ZeilerTF11}, 
and even crafting inputs to activate specific neurons~\cite{NguyenYC15}.
Recently, techniques such as \textit{Class Activation Mapping} (CAM)~\cite{ZhouKLOT16}, 
\textit{Gradient-weighted CAM} (Grad-CAM)~\cite{SelvarajuCDVPB17}, 
and Grad-CAM++~\cite{chattopadhay2018grad}, 
have pushed the boundaries. 
While CAM leverages global pooling to pinpoint informative regions, 
Grad-CAM offers a more general framework for visualizing any convolutional filter, 
revealing insights hidden within neural networks.
The quest for meaningful representations in face recognition goes back even further~\cite{ChenHLZ02,learned2016labeled,o2018face}. 
Traditional approaches often prioritized accuracy over interpretability, 
building representations from either facial parts~\cite{CaoYTS10,LiHLBY13} or facial attributes~\cite{KumarBBN09}. 
While these methods achieved progress, 
they lacked transparency in how they captured and utilized facial information, 
bridging the gap in the field of interpretable representation learning. 
% The field of interpretable representation learning aims to bridge this gap. 
% By delving deeper into the inner workings of models, 
% we can understand not just what, but also why and how they make decisions. 
% This newfound transparency holds immense potential for applications where trust and explainability are critical, 
% such as healthcare, law enforcement, and autonomous vehicles.

% Towards Interpretable Facial Emotion Recognition
% Towards Interpretable Face Recognition
\paragraph{Interpretable Emotion Classification.}

For centuries, the central pursuit in psychology has been studies of the kaleidoscope of human emotions.
In the 1960s, \citet{ekman1971constants} identified six basic universal expressions: 
happiness, sadness, anger, fear, surprise, and disgust. 
This framework later expanded to include contempt and embarrassment, 
offered a starting point for understanding emotional communication across cultures.
However, research has evolved, recognizing the more nuanced nature of emotions. 
Instead of discrete categories, 
recent models focus on ``emotional dimensions'': valence (positivity/negativity) and arousal (activation level). 
Proposed by \citet{russell1980circumplex}, 
these dimensions offer a broader spectrum, 
encompassing the vast range of human emotions beyond just eight facial expressions. 
This dimensional approach has found its way into numerous models of emotion recognition and synthesis, 
as seen in the work of \citet{KolliasTBCZ23} and \citet{tottenham2009nimstim}. 
Lately, 
\citet{YinTLS019} focused on a specific area of interpretable visual recognition by learning from data a structured facial representation. 
\citet{Malik0R21} proposed an interpretable deep learning-based system, 
which has been developed for facial emotion recognition tasks. 
% By embracing this more fluid understanding, 
% we gain a deeper insight into the complex tapestry of human emotions and their expression.

\section{Experimental Setup}
\label{sec:setup}

All the experiments are implemented in Python. 
We also use Shell and Jupyter Notebook for generating image and video scripts. 
The experiment and evaluation are conducted on two MacBook Pros 
(M1 Pro-Chip with 10-core CPU and 16-core GPU; Intel Core i9 with 2.3 GHz 8-Core). 

\begin{table*}[!ht]
  \centering
  \resizebox{.99\textwidth}{!}{
  \begin{tabular}{@{}lrcccccc|c@{}}
      \toprule
      \textsc{Dataset} & \textsc{Split} & \# happiness & \# surprise & \# sadness & \# anger & \# disgust & \# fear & \# total images/videos \\
      \midrule
      \multirow{2}{*}{RAF-DB~\cite{li_reliable_2017,li2019reliable}} & Train & 4772 & 1290 & 1982 & 705 & 717 & 281 & 9747 \\
      &Test &1185&329&478&162&160&74& 2388 \\
      \midrule
      \multirow{2}{*}{FER2013~\cite{BarsoumZCZ16}} & Train & 7215 & 3171 & 4830 & 3994 & 436 & 4097 & 23743 \\
      &Test &1774&831&1247&958&111&1024& 5945 \\
      \midrule
      FER AffectNet~\cite{Mollah2019ANet} & Train & 3091 & 4039 & 5044 & 3218 & 2477 & 3176 & 21045 \\
      \midrule
      CK+~\cite{LuceyCKSAM10} & Train & 69 & 83 & 28 & 45 & 59 & 25 & 309 \\
      \midrule
      TFEID~\cite{tfeid,LiGL22} & Train & 40 & 36 & 39 & 34 & 40 & 40 & 229 \\
      \midrule
      \midrule
      \multirow{3}{*}{FER \textsc{GiMeFive}} & Train  & 15187 & 8619 & 11923 & 7996 & 3729 & 7619 & \textbf{55073} \\
      & Test & 2959 & 1160 & 1725 & 1120 & 271 & 1098 & \textbf{8333} \\
      & Valid & 100 & 100 & 100 & 100 & 100 & 100 & \textbf{600} \\ 
      \bottomrule
      DISFA~\cite{MavadatiMBTC13} & Test & - & - & - & - & - & - & 54 \\
      \bottomrule
  \end{tabular}
  }
\caption{Overview of the data statistics for each emotion class and the total number of images/videos in our experiment 
(Note that - denotes the unknown information as the total videos are recorded from the participants randomly acting out their emotions).}
\label{tab:data}
\end{table*}

\subsection{Dataset}
\label{sec:setup:datasets}

% https://www.mdpi.com/1424-8220/22/21/8089

To initiate the project, 
we gathered image databases representing different types of emotional expressions from the
\textit{Real-world Affective Faces Database} (RAF-DB, in-the-wild expression)~\cite{li_reliable_2017,li2019reliable}, 
\textit{Facial Expression Recognition} 2013 (FER2013, real-time wild expression)~\cite{BarsoumZCZ16}, 
\textit{Facial Expression Recognition AffectNet Database} (FER AffectNet, in-the-wild expression)~\cite{Mollah2019ANet}, 
\textit{Extended Cohn-Kanade Dataset Plus} (CK+, posed expressions)~\cite{LuceyCKSAM10}, 
and \textit{Taiwanese Facial Expression Image Database} (TFEID, posed expressions)~\cite{tfeid,LiGL22}.
These image datasets come in folder-structure classification (See \cref{tab:data} 
for statistic details with specific numbers on training, test, and validation set for each emotion class). 

For reviewing explainable AI, 
we leverage the video dataset \textit{Denver Intensity of Spontaneous Facial Action Database} 
(DISFA, spontaneous expressions)~\cite{MavadatiMBTC13}, 
which comprises two sets of videos, 
differentiated by capturing various camera angles (left and right). 
Each set consists of 27 videos, with each video comprising 4844 frames, 
resulting in 130788 images for each camera angle and a total of 261576 images. 
The participants in the videos are performing a series of spontaneous facial actions. 
% which contains a variety of levels of intensity in expression. 

\paragraph{Image Preprocessing.}
The procedure is proceeded from public institutions and kaggle~\cite{kaggle_rafdb,kagaff}. 
To enhance the generalizability and robustness of our model, 
we aggregate these five FER image benchmarks into a new customized dataset called FER 
\textsc{GiMeFive}~\footnote{\url{https://github.com/werywjw/data}}. 
More specifically, 
the training set of FER \textsc{GiMeFive} is aggregated from the training sets of the five image datasets (55073 images in total), 
while the test set is combined from the test sets of RAF-DB and FER2013 (a total of 8333 images). 
The validation set is given with equal 100 images per class to test the performance of models. 

Building upon these image databases, we exclusively analyze human faces representing six emotions. 
That is, we first generalize a folder structure and append the emotion labels 0 to 5 to the name of each matching image 
(i.e., 0 for happiness, 1 for surprise, 2 for sadness, 3 for anger, 4 for disgust, and 5 for fear).
For implementation, 
we create a script for generating CSV files to store all the image names and their corresponding labels to later efficiently pass to the model. 
The images together with the CSV file are loaded and preprocessed for training. 
Therefore, we manipulate the pixel data through resizing, normalization, 
augmentation, and conversion to grayscale with three channels at $64 \times 64$ resolution in the JPG format, 
as our original CNN is designed to work with three-channel inputs. 
Typically, we assume that the color of the image does not affect the emotion classification. 

\paragraph{Video Preprocessing.}
For testing our model, we establish standards for extracted frames from videos or live webcam streams. 
Each frame, extracted for testing, undergoes scanning for the region of interest, the face(s). 
To achieve this, 
we implement the \texttt{CascadeClassifier}~\cite{casc_class} incorporating the Viola-Jones algorithm introduced by \citet{990517} to power, 
in our case, face detection, 
since the Viola-Jones algorithm revolutionizes real-time object detection. 
As part of the video preprocessing, 
we run every frame through the \texttt{FaceClassifier} to detect our region of interest and crop the rectangle region. 
The cropped image is also preprocessed by resizing to $64 \times 64$ resolution, 
then converted to greyscale with three channels, 
after that, a tensor is generated and normalized. 
Every cropped image is evaluated through our model so that the classification can be further used for interpretation. 

\subsection{Model Architecture}
\label{sec:setup:model}

\tikzstyle{block} = [rectangle, draw, fill=LMUGreen!30, text width=5em, text centered, rounded corners, minimum height=4em]
\tikzstyle{line} = [draw, -latex', line width=0.7pt]

\begin{figure*}[ht]
  \centering
  \resizebox{.99\textwidth}{!}{
  \begin{tikzpicture}[node distance=2.7cm, auto]
      \node [block] (input) {Input Layer};
      \node [block, right of=input] (conv1) {Conv Block 1 \\ 64 Filters};
      \node [block, right of=conv1] (conv2) {Conv Block 2 \\ 128 Filters};
      \node [block, right of=conv2] (conv3) {Conv Block 3 \\ 256 Filters};
      \node [block, right of=conv3] (conv4) {Conv Block 4 \\ 512 Filters};
      \node [block, right of=conv4] (conv5) {Conv Block 5 \\ 1024 Filters};
      \node [block, right of=conv5] (adaptivePool) {Adaptive Avg Pool};
      \node [block, right of=adaptivePool] (fc) {Fully Connected Block};
      \node [block, right of=fc] (softmax) {Softmax Layer};

      \path [line] (input) -- (conv1);
      \path [line] (conv1) -- (conv2);
      \path [line] (conv2) -- (conv3);
      \path [line] (conv3) -- (conv4);
      \path [line] (conv4) -- (conv5);
      \path [line] (conv5) -- (adaptivePool);
      \path [line] (adaptivePool) -- (fc);
      \path [line] (fc) -- (softmax);
  \end{tikzpicture}
  }
  \caption{Overview of the \textsc{GiMeFive} model architecture 
  (see \cref{fig:modeldetail} for a detailed version).} 
  \label{fig:model}
\end{figure*}
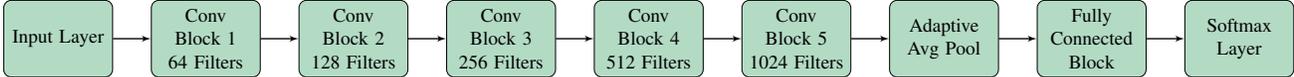

In order to compare the performance of our \textsc{GiMeFive} with other models, 
we replicate several state-of-the-art CNNs from scratch, 
which include \textit{Visual Geometry Group} (VGG)~\cite{SimonyanZ14a} and 
\textit{Residual Network} (ResNet)~\cite{HeZRS16} on two FER benchmarks (i.e., RAF-DB and FER 2013), 
as well as FER \textsc{GiMeFive} (see \cref{tab:data} for details). 

\paragraph{Visual Geometry Group.}
Developed by \citet{SimonyanZ14a}, 
the VGG model represents a milestone in the evolution of CNNs for image recognition tasks. 
Distinct for its simple architecture, 
the VGG employs a series of convolutional layers with small $3 \times 3$ convolution filters, 
followed by max-pooling layers, fully connected layers, and a softmax layer. 
This designed option allows the VGG to learn more complex patterns through the depth to 16-19 weight layers. 
We adapt the VGG by adding batch normalization and dropout to stabilize the learning process. 
Previous work has shown that the VGG model is effective in classifying large-scale facial emotions, 
as the representations generalize well to various datasets, 
where they achieve state-of-the-art results, 
confirming the importance of depth in visual representations. 

\paragraph{Residual Network.}
In principle, 
deeper neural networks with more parameters are more difficult to train. 
ResNet has emerged as a groundbreaking architecture, 
significantly advancing the performance on various tasks including image classification, 
object detection, and semantic segmentation. 
ResNets leverage the power of deep residual learning to achieve remarkable accuracies on benchmark datasets such as ImageNet. 

Inspired by \citet{HeZRS16}, 
we implement the ResNet18 and ResNet34 to ease the training process. 
ResNet18 and ResNet34 containing 18 and 34 layers respectively are specific configurations of the ResNet architecture, 
which address the vanishing/exploding gradient problem associated with training very deep neural networks through the introduction of residual blocks. 
These residual blocks allow for the direct propagation of gradients through the network by adding the input of the block to its output, 
effectively enabling the training of networks that are much deeper than previous architectures. 

\paragraph{\textsc{GiMeFive}.}
Our model architecture is illustrated in \Cref{fig:model} (see \cref{fig:modeldetail} for details). 
The input of our emotion recognition model is an image with 3 channels at $64 \times 64 $ resolution. 
The output is six emotion classes. 
% , i.e., happiness, surprise, sadness, anger, disgust, and fear. 
We implement an emotion classifier from scratch with four convolution blocks as our baseline at the very beginning. 
Despite the larger kernel being able to provide more information and a wider area view due to more parameters, 
we use a $3 \times 3$ kernel size for all convolutional layers, 
which is the same as the VGG model, 
since it is efficient to train and share the weights without expensive computation. 
Following each convolutional layer, 
batch normalization is used for stabilizing the learning by normalizing the input to each layer. 
Also, 
the batch normalization ensures forward propagated signals have non-zero variances. 

The convolution stride is fixed to 1 pixel. 
We interleave with the max pooling layer because it reduces the spatial dimensions of the input volume. 
Afterward, three linear layers are applied to extract features to the final output. 
We also add a 50\% dropout layer between the first and second linear layers to prevent overfitting. 
The activation function after each layer is \textit{Rectified Linear Unit} (ReLU), 
since it introduces the non-linearity into the model, 
allowing it to learn more complex patterns. 
Note that our best model \textsc{GiMeFive} is later optimized with the dropout rate at 20\% after max pooling layers. 

The softmax function plays a critical role in the final stage of a neural network, 
which corresponds to different classes by exponentiating each logit and then normalizing these values by the sum of all exponentiated logits. 
Mathematically, for a given logit $z_i$ among $n$ classes, 
the softmax function $\mathcal{S}(z_i)$ is defined as: 
\begin{equation}
  \label{eq:softmax}
  \mathcal{S}(z_i) = \frac{e^{z_i}}{\sum_{j=1}^{n} e^{z_j}} \in [0,1], 
\end{equation}
where the true class has probability 1 and all others have probability 0. 

\begin{table*}[ht]
  \centering
  % \resizebox{.99\textwidth}{!}{
  \begin{tabular}{@{}llcccccr@{}}
    \toprule 
    \multirow{2}{*}{\textsc{Dataset}}&\multirow{2}{*}{\textsc{Models}}&\multirow{2}{*}{\textsc{\# Layers}}&\multirow{2}{*}{\textsc{Architecture}} & \multicolumn{3}{c}{\textsc{Accuracies}} & \multirow{2}{*}{\textsc{\# Parameters}} \\
    \cline{5-7}
    &&&& Train & Test & Valid  &  \\
    \midrule
    \multirow{8}{*}{RAF-DB~\cite{li_reliable_2017,li2019reliable}} &VGG~\cite{SimonyanZ14a} & 16 &+BN&93.8&83.3&68.8&72460742\\ 
    & ResNet18~\cite{HeZRS16} & 18 & +RB  & 98.9 & 81.3 & 67.9 & 11179590 \\
    & ResNet34~\cite{HeZRS16} & 34 & +RB  & \textbf{100.0} & 83.7 & 72.0 & 21287750 \\
    &\textsc{GiMeFive} (Baseline) & 13 & +BN-SE & 96.6 & 80.6 & 66.8 & 2606086 \\ 
    % &\textsc{GiMeFive} (Ours) & 10 & -BN-SE & 96.3 & 76.9 & 60.6 & 10474118 \\
    &\textsc{GiMeFive} (Ours) & 16 & +BN+SE & 98.4 & 81.7 & 71.1 & 10478598 \\
    &\textsc{GiMeFive} (Ours) & 15 & +BN-SE & 98.6 & 83.1 & 72.1 & 10478086 \\
    &\textsc{GiMeFive} (Ours)\textcolor{LMUGreen}{$^\star$} & 15 & +DO+BN-SE & 97.0 & \textbf{86.5} & \textbf{78.1} & 10478086 \\
    &\textsc{GiMeFive} (Ours) & 17 & +BN-SE & 97.5 & 82.5 & 70.0 & 41950726 \\ 
    \midrule
    \multirow{7}{*}{FER2013~\cite{BarsoumZCZ16}} & VGG~\cite{SimonyanZ14a} & 16 & +BN & 69.0 & 49.4 & 35.6 & 72460742 \\
    & ResNet18~\cite{HeZRS16} & 18 & +RB  & 92.6 & 64.7 & 42.1 & 11179590 \\
    & ResNet34~\cite{HeZRS16} & 34 & +RB  & \textbf{97.6} & 67.1 & 43.6 & 21287750 \\
    & \textsc{GiMeFive} (Baseline) & 13 & +BN-SE & 86.6 & 64.1 & 40.2 & 2606086 \\
    &\textsc{GiMeFive} (Ours) & 15 & +BN-SE & 89.6 & 65.6 & 40.7 & 10478086 \\
    &\textsc{GiMeFive} (Ours)\textcolor{LMUGreen}{$^\star$} & 15 & +DO+BN-SE & 69.7 & \textbf{67.5} & \textbf{46.2} & 10478086 \\
    &\textsc{GiMeFive} (Ours) & 17 & +BN-SE & 96.0 & 65.5 & 41.6 & 41950726 \\
    \midrule
    \midrule
    \multirow{5}{*}{FER \textsc{GiMeFive}} & VGG~\cite{SimonyanZ14a} & 16 & +BN & 79.4 & 53.7 & 61.4 & 72460742 \\
    & ResNet18~\cite{HeZRS16} & 18 & +RB  & \textbf{96.5} & 72.4 & 73.8 & 11179590 \\
    & ResNet34~\cite{HeZRS16} & 34 & +RB  & 94.5 & 70.8 & 72.8 & 21287750 \\
    &\textsc{GiMeFive} (Ours)\textcolor{LMUGreen}{$^\star$} & 15 & +DO+BN-SE & 84.9 & \textbf{75.3} & \textbf{75.0} & 10478086 \\
    &\textsc{GiMeFive} (Ours) & 25 & +BN+SE+RB & 95.2 & 70.4 & 74.5 & 95051014 \\
    \bottomrule
  \end{tabular}
  % }
  \caption{Accuracies (\%) for different models with specific architectures and numbers of parameters and layers in our experiments 
  (Note that \textcolor{LMUGreen}{$^\star$} denotes our best GiMeFive; 
  BN stands for the batch normalization, 
  RB for residual block, 
  SE for the squeeze and excitation block, 
  and DO for dropout; 
  +/- represent with/without respectively). 
  The confusion matrix of the validation set in terms of the number of classified images is shown in \cref{fig:matval}, 
  while the result of the test set on the RAF-DB is given in \cref{fig:mattest}.} 
  \label{tab:model}
\end{table*}

\section{Evaluation}
\label{sec:evaluation}

For evaluation, we use the metric accuracy to see if our model can classify facial emotions correctly. 
We report all the training, testing, and validation accuracies in \% 
to compare the performance of our \textsc{GiMeFive} with other state-of-the-art methods. 
By converting logits to probabilities, 
the softmax function allows for a more interpretable output from the model and facilitates the use of probability-based loss functions. 
The loss function employed for all models is \textit{cross-entropy} (CE), 
which is typically for multi-class classification during training. 
The CE loss is given by: 
\begin{equation}
  \label{eq:ce}
  \mathcal{L}_{\text{CE}} = -\sum_{i=1}^{n} y_i \log(p_i),
\end{equation}
where $y_i$ is the true label and $p_i$ is the predicted probability of the $i$-th class. 
Here $n$ denotes the total number of classes, in our case, 6.

\subsection{Evaluation Results}
\label{sec:evaluation:results}

As seen in \Cref{tab:model}, 
the results in terms of accuracies are aggregated from the database RAF-DB~\cite{kaggle_rafdb} and FER2013~\cite{kaggle_fer}, 
as well as the FER \textsc{GiMeFive}. 
We observe that our models achieve the best performance on RAF-DB instead of on our aggregated FER \textsc{GiMeFive}. 
We assume that the reason for this is that the RAF-DB dataset is more balanced, 
while the FER2013 dataset is more challenging due to the imbalance of the classes. 
Also verified on the test and validation set, 
the performance of all our models on FER2013 is worse than on the RAF-DB. 

At the same time, 
we can see that there is a trade-off between accuracy and the number of parameters. 
In general, 
deeper CNNs with more layers and parameters tend to perform better, 
as they can capture more nuanced facial patterns and features. 
However, in our experiments, 
this is not always the case. 
We hypothesize that more parameters often mean greater learning capacity, 
but also a higher risk of overfitting and increased computational cost. 
The models are then too complex and need more resources and time to train. 
Therefore, we propose our best model \textsc{GiMeFive}, 
the one with 15 layers and 10478086 parameters, 
which is seven times less than the VGG and half of ResNet34, 
efficiently achieving the best performance on the both test and validation set on all benchmarks. 

In terms of the implementation details, 
our \textsc{GiMeFive} 
(marked as a green asterisk \textcolor{LMUGreen}{$\star$}) without random augmentation outperforms the other methods in terms of accuracy, 
indicating that this kind of random augmentation is not necessarily able to help our model predict the correct label, 
one reason is that the images on RAF-DB are already aligned and cropped, 
the \texttt{RandomHorizontalFlip}, \texttt{RandomRotation}, 
\texttt{RandomCrop}, and \texttt{RandomErasing} might confuse the model by adding more noise to aligned facial images. 
We further continue to simply apply merely normalization to detect more representative features with respect to different emotions.
Previous research from \citet{ZeilerF14,li_reliable_2017,VermaMRMV23} engage similar investigations. % for simplicity. 

We run extensive experiments to determine the number of convolutional blocks for a better trade-off between the number of parameters and the performance. 
Eventually, we find that adding an extra convolutional block 
(5 blocks with 15 layers in total) from our initial baseline (4 blocks with 13 layers)
leads to the best performance. 
However, six convolutional blocks with 17 layers do not necessarily result in better performance, 
this is due to the complexity and overfitting of the models. 

Moreover, 
we find that the dropout and batch normalization can indeed improve the performance of the model, 
as such regularization techniques can prevent overfitting by randomly dropping units 
(along with their connections) from the neural networks during training. 
Also, % stabilize the learning process and
batch normalization can help the model reduce the number of training epochs, 
leading to faster learning progress and thus improving the generalization ability of the model. 
Unfortunately, the squeeze and excitation block~\cite{HuSASW20}, 
a mechanism that adaptively recalibrates channel-wise attention feature responses by explicitly modeling interdependencies between channels, 
did not improve the performance in this task. 
One possible interpretation is that emotion classification might benefit more from spatial relationships rather than channel-wise dependencies. 

Notably, 
ResNet18 is appreciated for its efficiency and lower computational cost, 
and ResNet34 for offering a balance between depth and performance, 
making them widely adopted in both academic research and practical applications, 
achieving good performance on two FER benchmarks in our empirical experiment as well. 
However, due to the lack of further optimization techniques, % such as hyperparameter searching,  on these state-of-the-art models, 
they did not significantly outperform our best model \textsc{GiMeFive}. 
We agree that the innovative approach of using residual connections has not only improved the performance of deep neural networks 
but also inspired a plethora of subsequent research efforts for informative facial feature detections. % aimed at enhancing network architectures

\subsection{Interpretable Results}
\label{sec:evaluation:inter}

\begin{figure}[ht]
  \centering
   \includegraphics[width=\linewidth]{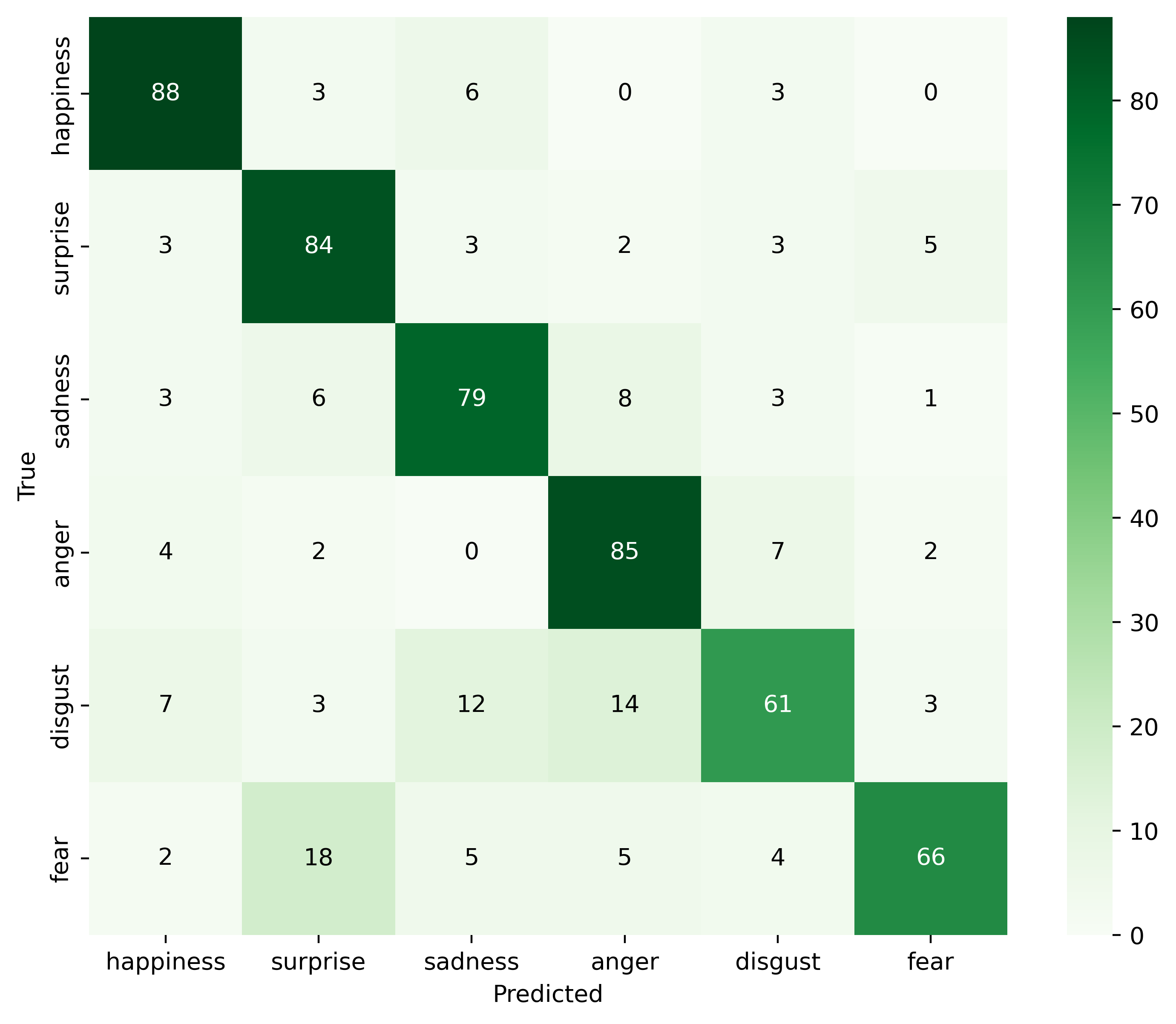}
   \caption{Overview of the confusion matrix evaluated from our \textsc{GiMeFive} on the validation set.} 
   \label{fig:matval}
\end{figure}

\paragraph{Classification Scores.} 
To further analyze the probability of each individual image, 
we write a script that takes a folder path as input and iterates through the images inside a subfolder to record the performance of our best model \textsc{GiMeFive}. 
This CSV file is represented with the corresponding classification scores via the softmax function (see \cref{eq:softmax}). 
To have a better review of the total correctly classified images, 
we also generate a confusion matrix heatmap (see \cref{fig:matval}) to visualize the performance of the model with respect to each emotional class, 
where the numbers in the diagonal of this heatmap represent the number of correct predicted images on the validation set. 
Each line has a sum of 100, 
as our validation set contains one hundred images with the ground-truth label per class (see \cref{tab:data}). 

As shown in \Cref{fig:matval}, 
we observe that our \textsc{GiMeFive} can predict the happiness class with the highest accuracy, 
while the class with the lowest accuracy is disgust. 
This is understandable since facial emotions such as sadness and anger are similar to disgust, 
especially when humans sometimes also represent disgusting and angry emotions at the same time. 
Roughly speaking, 
happiness is the largest sub-class in our training data (see \cref{tab:data}) and the happy emotion is the easiest to detect in principle. 
Moreover, 
the confusion matrix illustrates that the model has good performances in terms of classifying surprise, sadness, and anger as well. 

\paragraph{Overlay Grad-CAM.}
To understand the decision-making process of our model, 
we aim to explain our model in a more transparent and interpretable way using Grad-CAM~\cite{SelvarajuCDVPB17}, 
as Grad-CAM is a generalization of CAM and helps interpret CNN decisions 
by providing visual cues about the regions that influenced the classification. 
Grad-CAM highlights the important regions of an image, 
aiding in the understanding of the behavior of the model, 
which is especially useful for model debugging and further improvement. 

\begin{figure}[ht]
  \centering
  \begin{subfigure}{0.47\linewidth}
    \includegraphics[width=\linewidth]{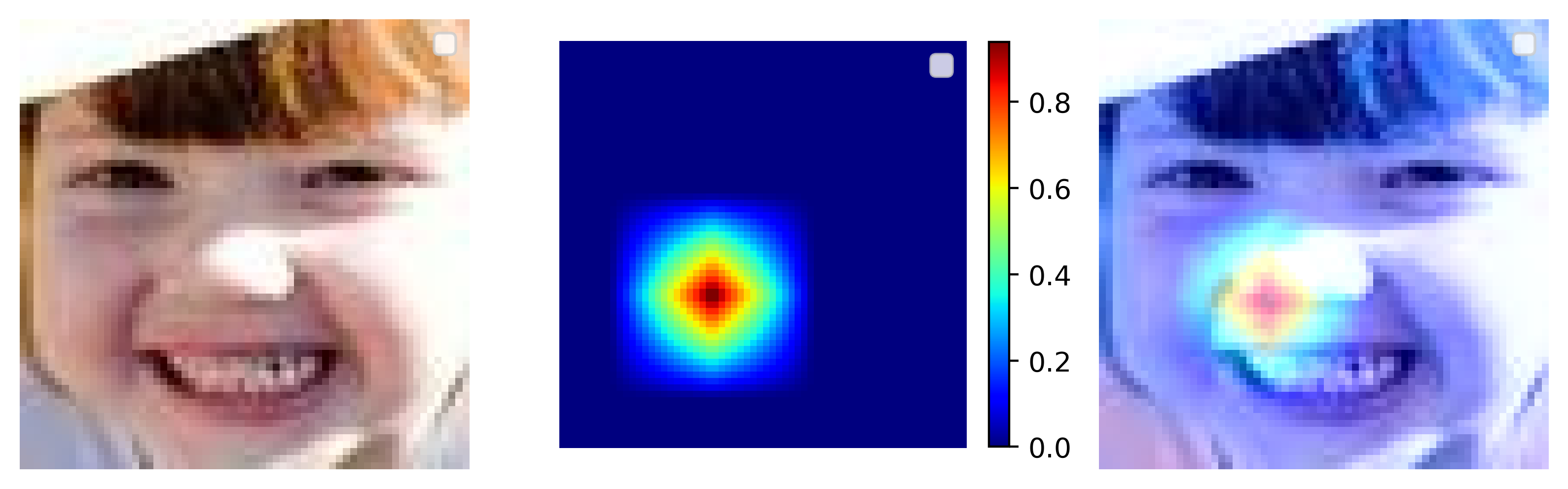}
    \caption{Happiness.}
    \label{fig:xai1}
  \end{subfigure}
  \hfill
  \begin{subfigure}{0.47\linewidth}
    \includegraphics[width=\linewidth]{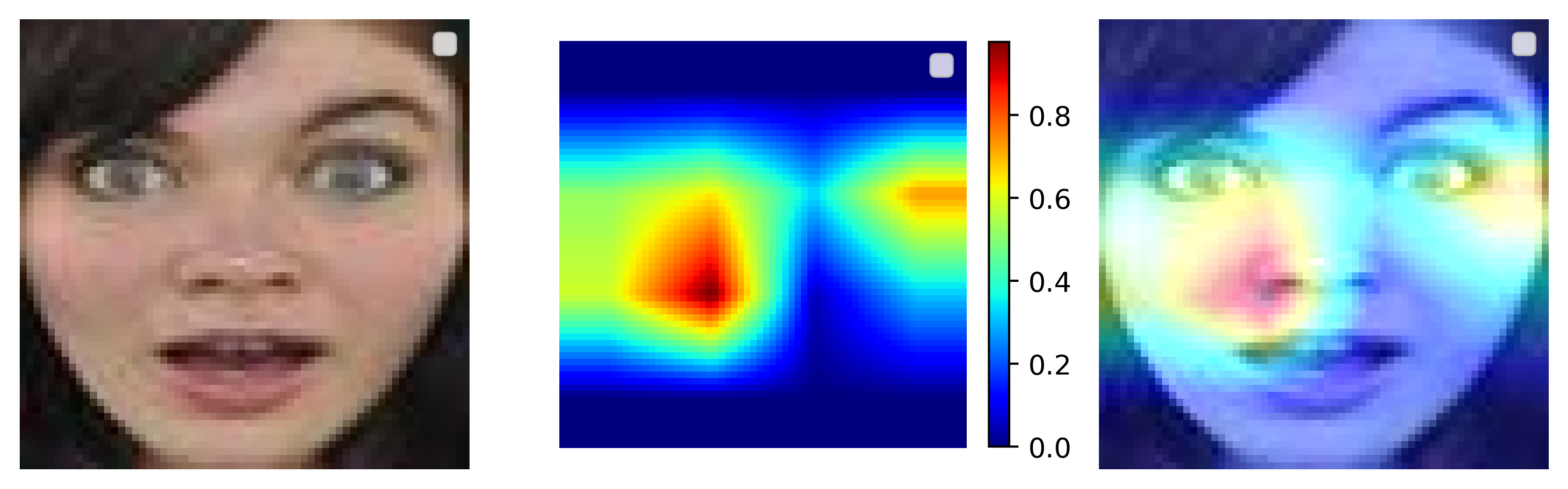}
    \caption{Surprise.}
    \label{fig:xai2}
  \end{subfigure}
  \hfill
  \begin{subfigure}{0.47\linewidth}
    \includegraphics[width=\linewidth]{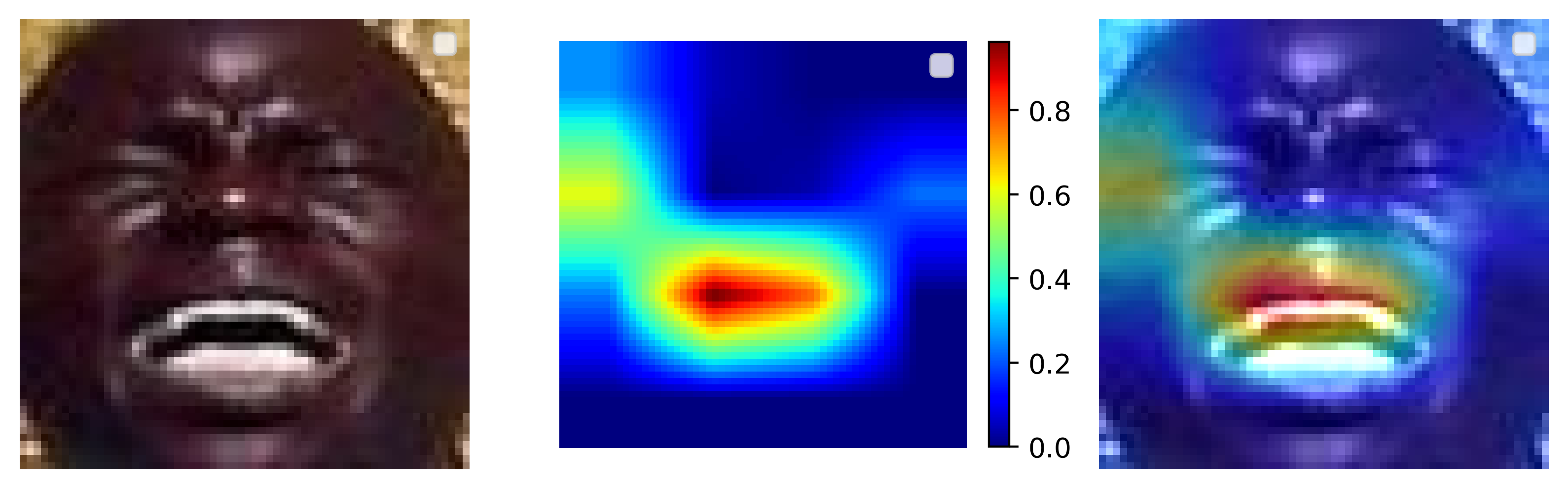}
    \caption{Sadness.}
    \label{fig:xai3}
  \end{subfigure}
  \hfill
  \begin{subfigure}{0.47\linewidth}
    \includegraphics[width=\linewidth]{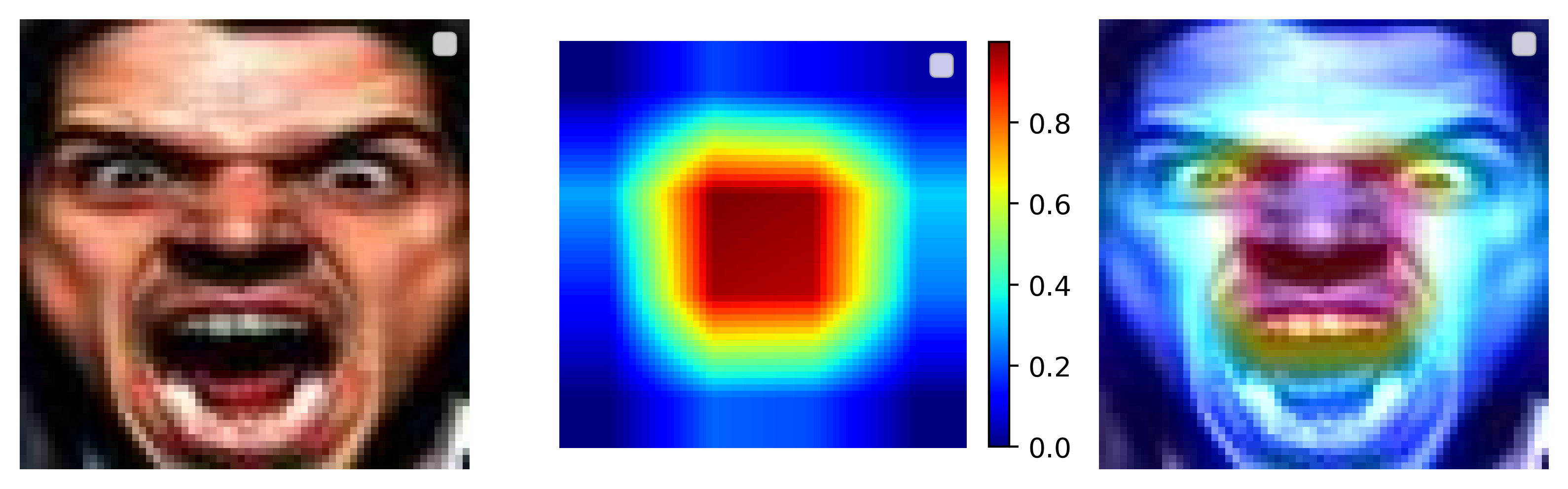}
    \caption{Anger.}
    \label{fig:xai4}
  \end{subfigure}
  \hfill
  \begin{subfigure}{0.47\linewidth}
    \includegraphics[width=\linewidth]{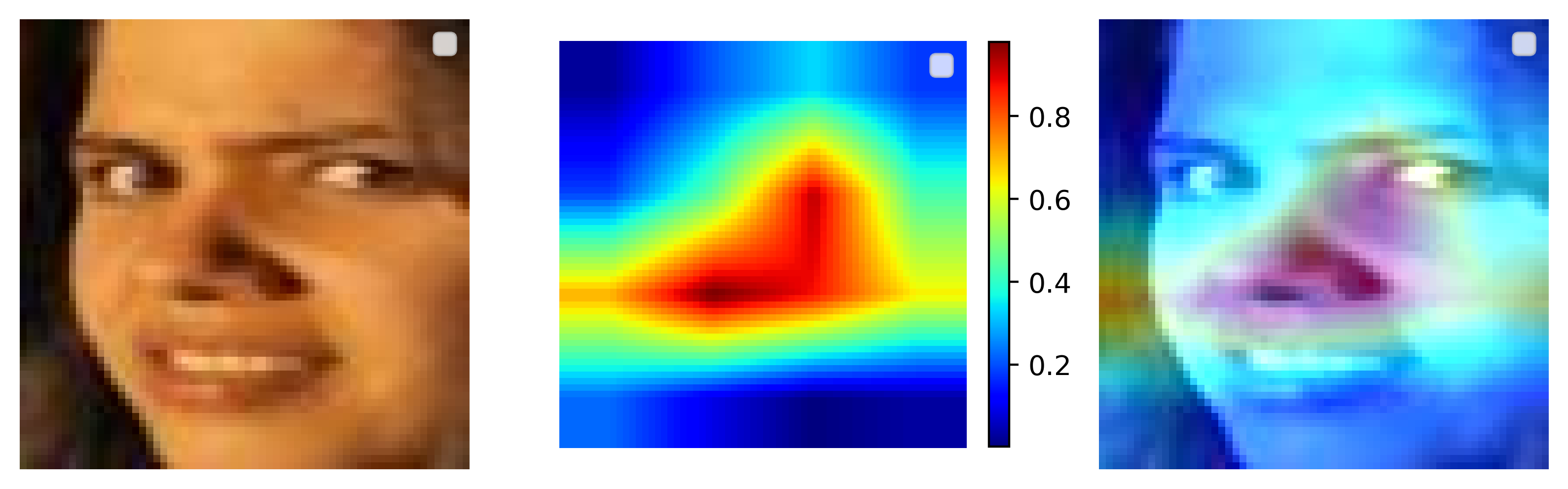}
    \caption{Disgust.}
    \label{fig:xai5}
  \end{subfigure}
  \hfill
  \begin{subfigure}{0.47\linewidth}
    \includegraphics[width=\linewidth]{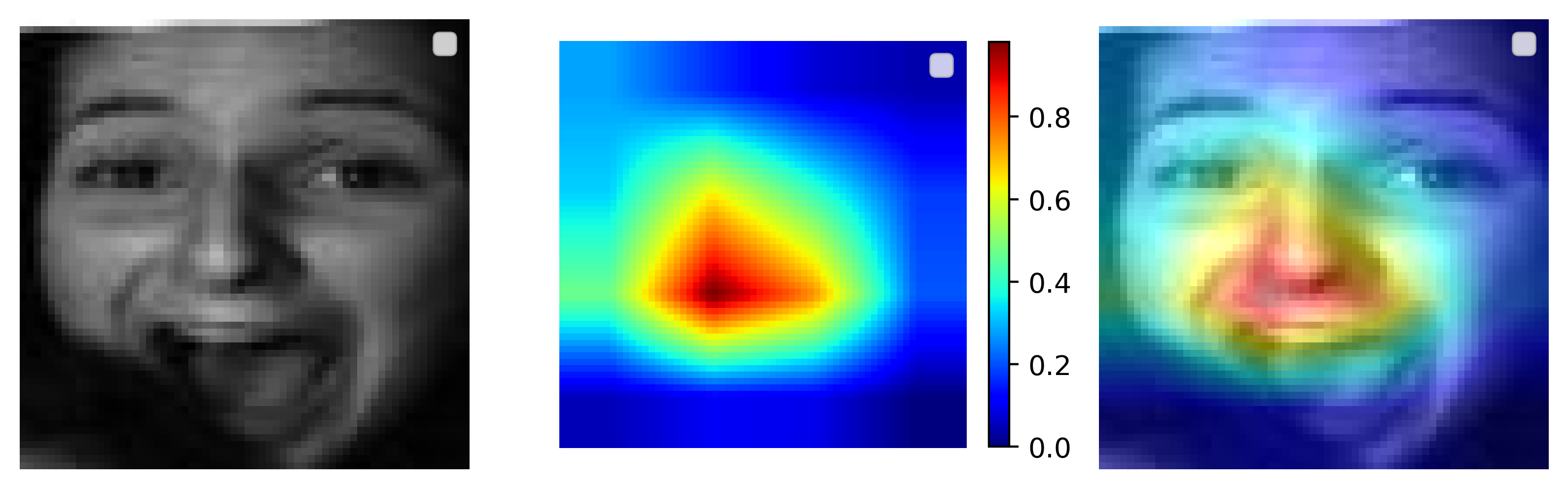}
    \caption{Fear.}
    \label{fig:xai6}
  \end{subfigure}
  \caption{Interpreting images with all six facial emotional classes (left: original image; middle: heatmap; right: Grad-CAM) evaluated from our \textsc{GiMeFive} on the validation set.}
  \label{fig:xai}
\end{figure}

Besides proposing a method to visualize the discriminative regions of a CNN trained for the classification task, 
we adopt this approach from \citet{ZhouKLOT16} to localize objects without providing the model with any bounding box annotations. 
The model can therefore learn the classification task with class labels and is then able to localize the object of a specific class in an image. 
In our case, we leverage the last (i.e., fifth) convolutional layer of our \textsc{GiMeFive} to generate the Grad-CAM heatmap. 
The \textit{global average pooling} 
(GAP) layer is utilized to compute the average value of each feature map, 
resulting in a spatial average of feature maps.

As seen in \Cref{fig:xai}, 
the heatmap in the middle represents the results of our best model \textsc{GiMeFive} on the validation set, 
showing different areas of the face to identify each emotion. 
The warmer colors (red and yellow) indicate the higher activation or relevance in recognizing facial emotions, 
while the cooler colors (blue and green) indicate the lower activation. 
The scale ranges from 0 to 1, 
indicating the level of activation or confidence in emotion recognition,
where 1 means the highest measured activation. 

The right side of each sub-figure is a result of the overlay Grad-CAM. 
For instance, 
we can see that the happiness and sadness (see \cref{fig:xai1} and \cref{fig:xai3} respectively) 
classes show a high level of activation around the mouth region, 
which is common in uplifting and drooping corners. 
Surprise (see \cref{fig:xai2}) implies activation around the eyes and eyebrows, which are typically raised in a look in human faces. 
Other emotions such as anger, disgust, and fear (see \cref{fig:xai4}, \cref{fig:xai5}, and \cref{fig:xai6} respectively) 
represent entire distributions, highlighting the complexity of facial expressions from our models.

\begin{figure*}[ht]
  \centering
  \begin{subfigure}{0.49\linewidth} % 0.24
    \includegraphics[width=\linewidth]{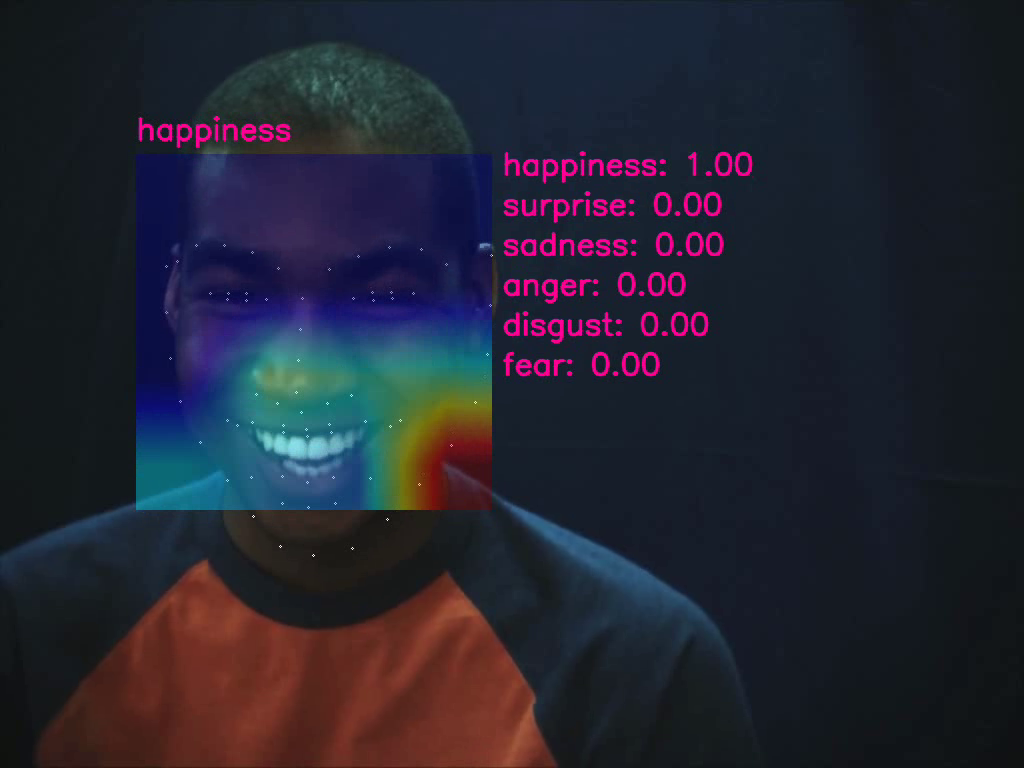}
    \caption{Happiness.}
    \label{fig:v1}
  \end{subfigure}
  \hfill
  \begin{subfigure}{0.49\linewidth}
    \includegraphics[width=\linewidth]{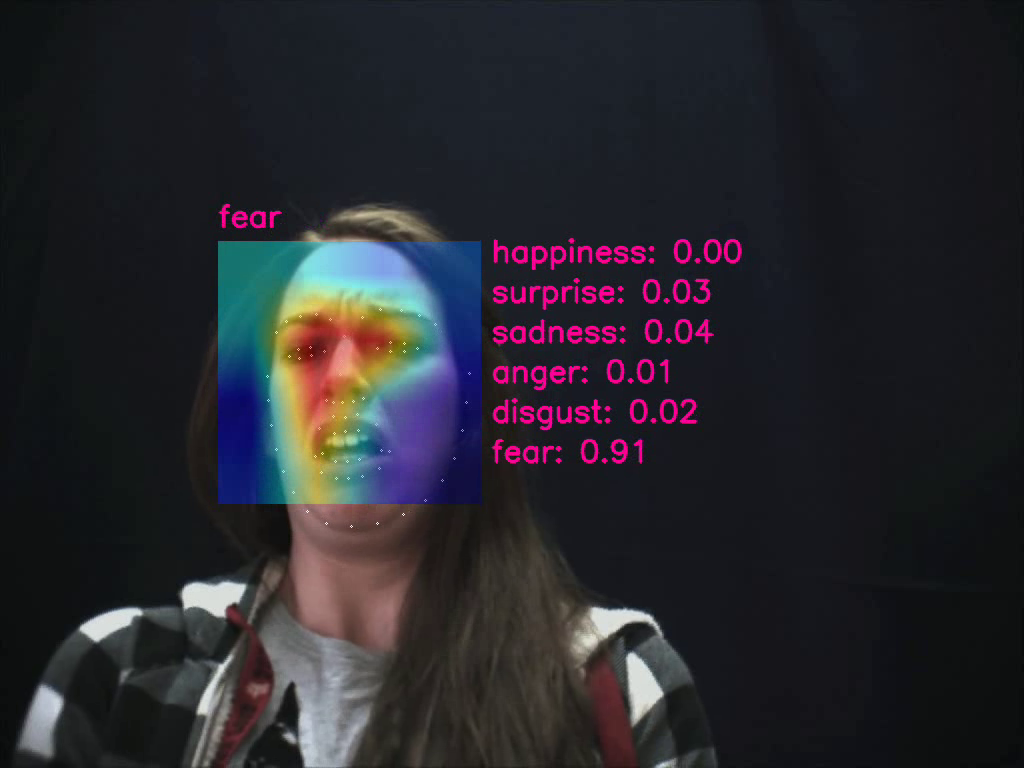}
    \caption{Fear.}
    \label{fig:v2}
  \end{subfigure}
  \hfill
  \begin{subfigure}{0.49\linewidth}
    \includegraphics[width=\linewidth]{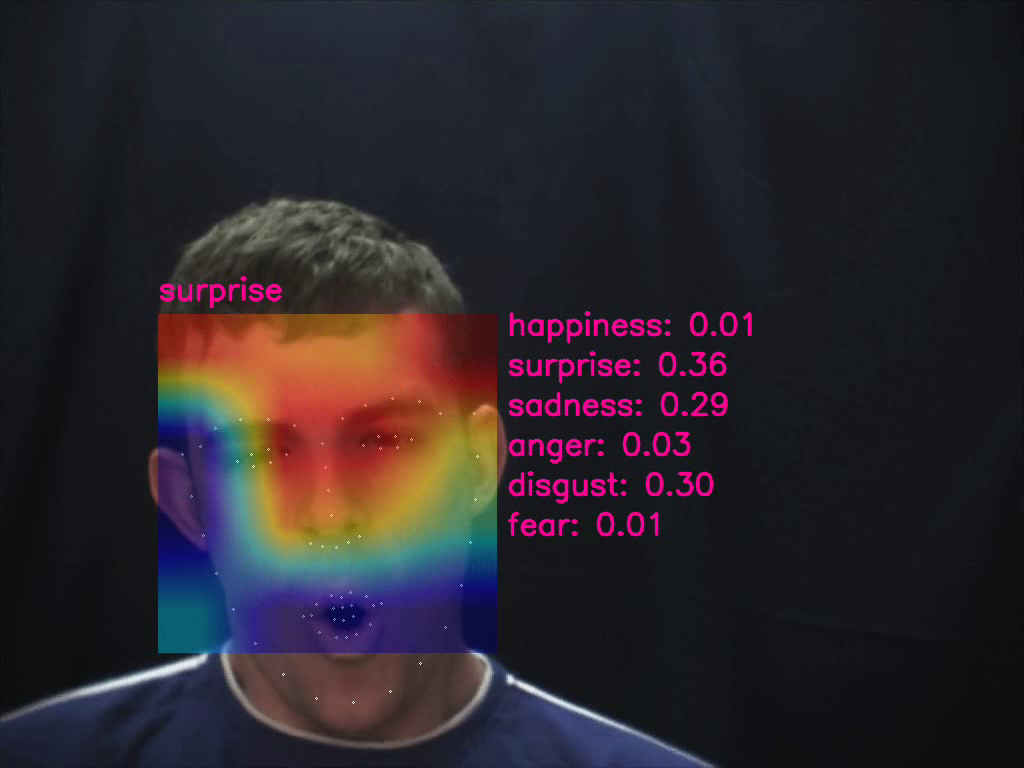}
    \caption{Surprise.}
    \label{fig:v3}
  \end{subfigure}
  \hfill
  \begin{subfigure}{0.49\linewidth}
    \includegraphics[width=\linewidth]{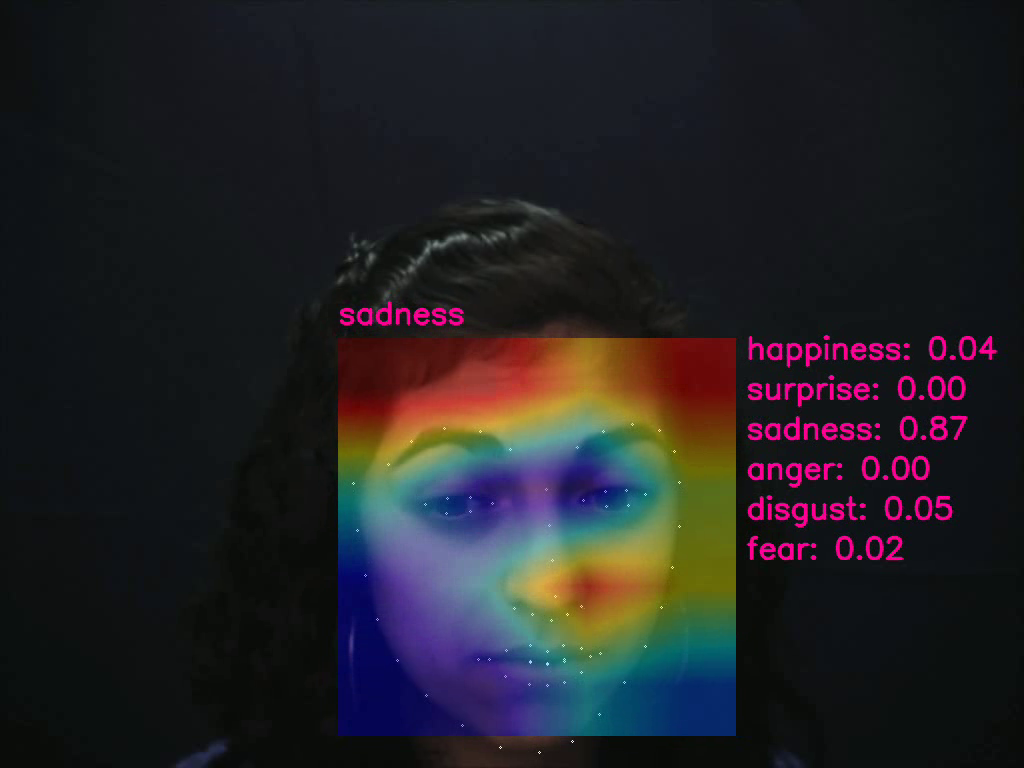}
    \caption{Sadness.}
    \label{fig:v4}
  \end{subfigure}
  % \hfill
  % \begin{subfigure}{0.32\linewidth}
  %   \includegraphics[width=\linewidth]{video/v_disgust.png}
  %   \caption{Sadness.}
  %   \label{fig:v5}
  % \end{subfigure}
  % \hfill
  % \begin{subfigure}{0.32\linewidth}
  %   \includegraphics[width=\linewidth]{video/v_fear.png}
  %   \caption{Sadness.}
  %   \label{fig:v6}
  % \end{subfigure}
  % \hfill
  \caption{Interpreting video screenshots with grad-CAM and landmarks evaluated from our \textsc{GiMeFive} on DISFA dataset.}
  \label{fig:video}
\end{figure*}

\paragraph{Overlay Landmark.}
The \textsc{GiMeFive} classifications of every frame are plotted in percentage next to the saliency map~\cite{SimonyanVZ13}.
The most important emotion, extracted from every fifth frame, 
is displayed above the map to make it easier for our eyes to follow. 
To plot landmarks on each video frame, 
we implemented the pre-trained \texttt{dlib shape predictor}, \texttt{shape predictor 68 face landmarks.dat}, 
constructed using the classic \textit{Histogram of Oriented Gradients} (HOG)~\cite{1467360} feature combined with a linear classifier, 
an image pyramid, and a sliding window detection scheme~\cite{dlib_site}. 
The pose estimator was created using the \texttt{dlib} from \citet{6909637}.

With this pre-trained model, 
68 numbered landmarks assist the emotion analysis and facial expression tracking to understand and interpret facial features.
% We implement using the libraries such as Dlib~\cite{dlib_site}. 
Again, illustrated in \Cref{fig:video}, 
happiness emotions (see \Cref{{fig:v1}}) are the easiest to detect in principle. 

\section{Optimization Strategies}
\label{sec:optim}

\paragraph{Data Augmentation.}
In deep learning and AI, 
data augmentation~\cite{augment} stands as a transformative technique, 
empowering algorithms to learn from and adapt to a wider range of data. 
This is typically beneficial when the dataset is limited, 
by introducing subtle modifications to existing data points, 
augmentation effectively expands the dataset, 
enabling models to generalize better and achieve enhanced performance. 

Preliminary, 
we create various replications of existing photos by randomly altering different properties such as size, 
brightness, color channels, or perspectives on the RAF-DB dataset. 
The results, however, are not as expected since aligned and cropped images do not have significant advantages in helping our \textsc{GiMeFive}. 
In the future, 
we believe this technique can be more effective when applied to the other state-of-the-art models on real-world variations, 
which can slightly encounter altered versions of familiar data, 
making more nuanced and robust predictions. 
% Additionally, we guide the training process to enhance the recognition and handling of real-world variations.
% Different combinations of functions from the \texttt{pytorch.transforms} library are tested for augmentation from those already established filters. 

\paragraph{Hyperparameter Tuning.}
To understand and enhance the performance of the model during training 
and find the best hyperparameter configuration (see \cref{tab:hyper} for details) of the model, 
we utilize the \texttt{parameter grid}~\cite{parameterGrid}. 
As a result, 
The learning rate for all our \textsc{GiMeFive} models is set to $0.001$ with the stochastic gradient descent optimizer. 
The \texttt{momentum} $=0.9$, 
meanwhile, we apply the \texttt{weight decay} $ = \num{1e-4}$. 
Additionally, 
we increase the depth of the network by adding some convolutional layers to learn more complex features. 
To help the training of deeper neural networks more efficiently, 
we add the residual connections, 
as they allow gradients to flow through the network more easily, 
improving the training for deep architectures. 
As verified by \citet{BarsoumZCZ16}, 
the dropout layers are effective in avoiding model overfitting. 
Potential, 
more advanced optimization techniques such as random search, bayesian optimization, 
and genetic algorithms are also able to be further explored to find the best hyperparameters for our \textsc{GiMeFive}.

\begin{table}[ht]
  \centering
  \begin{tabular}{@{}lc@{}}
    \toprule
    \textsc{Hyperparameter} & \textsc{Value} \\
    \midrule
    Learning rate & $ \{\num{1e-2}, \num{1e-3}, \num{1e-4} \} $ \\
    Batch size & \{8, 16, 32, 64\} \\
    Dropout rate & \{0.2, 0.5\} \\
    Convolution depth & \{4, 5, 6\} \\
    Fully connected depth & \{1, 2, 3\} \\
    Batch normalization & \{\texttt{True}, \texttt{False}\} \\
    Pooling & \{\texttt{max}, \texttt{adaptive avg}\} \\
    Optimizer & \{\texttt{Adam}, \texttt{AdamW}, \texttt{SGD}\} \\
    Activation & \{ \texttt{relu}\} \\ % , \texttt{tanh}, \texttt{elu}, \texttt{gelu}
    Epoch & \{10, 20, 40, 80\} \\
    Early stopping & \{\texttt{True}, \texttt{False}\} \\
    Patience & \{5, 10, 15\} \\
    \bottomrule
  \end{tabular}
  \caption{Explored hyperparameter space for our models.}
  \label{tab:hyper}
\end{table}

\section{Conclusion and Discussion}
\label{sec:conclusion}

In this work, we propose a novel emotion classifier \textsc{GiMeFive} for facial expression analysis, 
offering novel insights and methodologies for enhancing the interpretability and accuracy of facial emotion classifications. 
We run extensive experiments on two FER benchmarks and a five-dataset aggregated FER \textsc{GiMeFive} to evaluate the performance of our models, 
finding that our best model \textsc{GiMeFive} outperforms other state-of-the-art models with both efficiency and explainability. 
% with 15 layers and 10478086 parameters 
In addition, 
we also provide several demonstration videos and live web camera streams to illustrate the interpretability of our model. 
We hope that our work can inspire future research in the field of facial emotion classification and multimodal analysis. 

\paragraph{Limitations.}
We acknowledge that our work has several constraints in terms of the accuracy of the models on the leaderboard. 
The limited time frame allocated for the study might have influenced both the depth and the breadth of the investigation. 
We believe that the performance of our model can be further improved by incorporating more advanced optimization techniques 
and loss functions. 
% ~\cite{LiGL22}. 
% Please feel free to contact us if you have any questions or further suggestions.

\section*{Acknowledgements}

We are deeply grateful to our advisors \textbf{Johannes Fischer} and \textbf{Ming Gui} for their helpful feedback and valuable support during the entire semester. 
We also thank \textbf{Prof. Dr. Björn Ommer} for providing this interesting practical course. 
A special acknowledgement to our former team members \textbf{Mahdi Mohammadi}, who joined us till the end of the 
second phase of the project and shared research in conclusion, data pre-processing, and CAM-Images inquiry, 
and \textbf{Tanja Jaschkowitz}, who joined us during the first phase of the project and shared two Jupyter Notebook scripts.

%% file: sec/appendix.tex
\begin{figure}[ht]
  \centering
   \includegraphics[width=\linewidth]{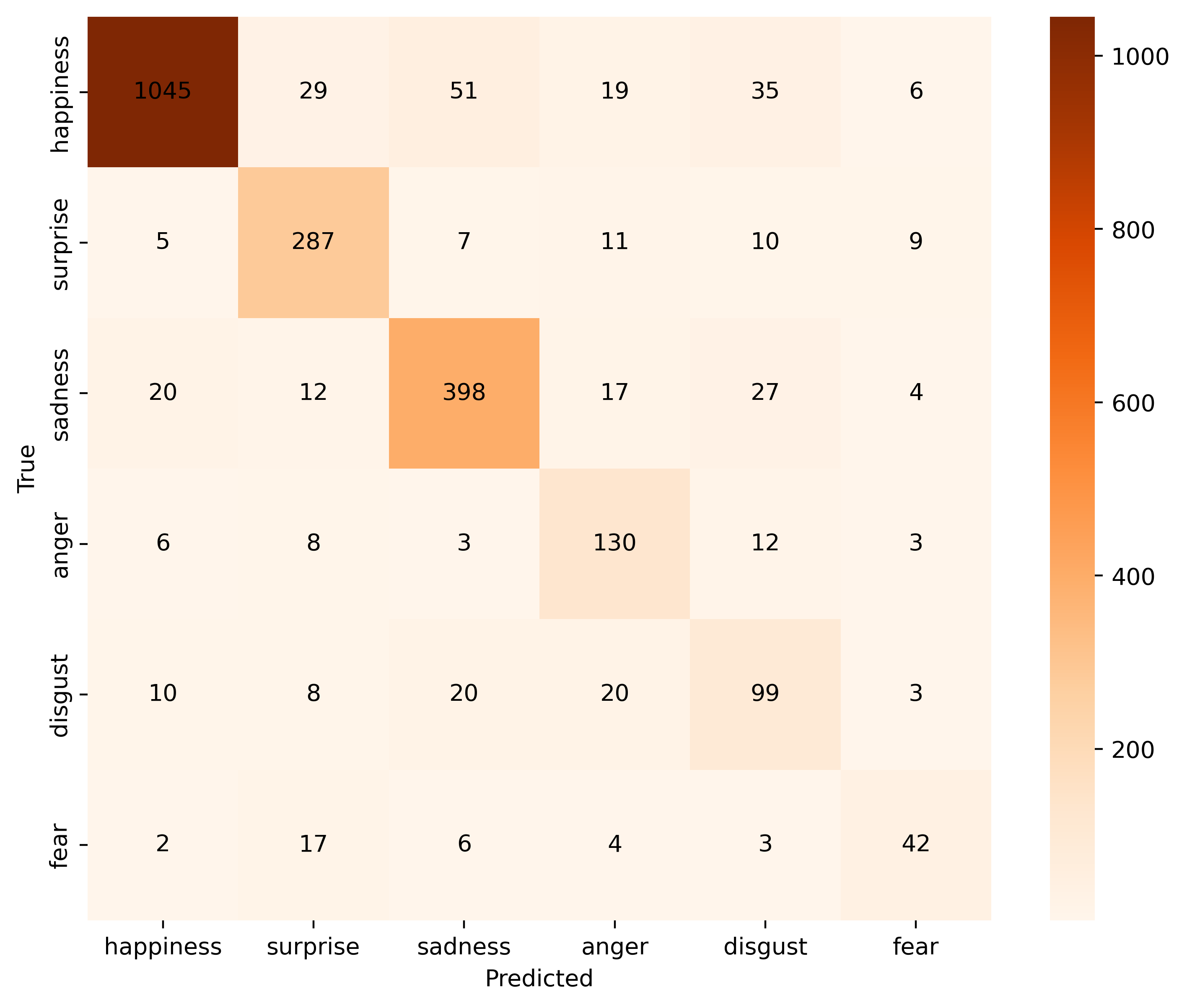}
   \caption{Overview of the confusion matrix on the test set of RAF-DB.} 
   \label{fig:mattest}
\end{figure}

\tikzstyle{layer} = [rectangle, rounded corners, minimum width=3cm, minimum height=0.8cm,text centered, draw=black, fill=LightPurple!30]
\tikzstyle{pool} = [rectangle, rounded corners, minimum width=3cm, minimum height=0.8cm, text centered, draw=black, fill=cvprblue!30]
\tikzstyle{fc} = [rectangle, minimum width=3cm, minimum height=0.8cm, text centered, draw=black, fill=Orange!30]
\tikzstyle{do} = [rectangle, minimum width=3cm, minimum height=0.8cm, text centered, draw=black, fill=LMUGrey2!30]
\tikzstyle{arrow} = [thick,->,>=stealth]
% \tikzset{arrow/.style={thick,-{Stealth[length=10mm,width=2mm]}}}

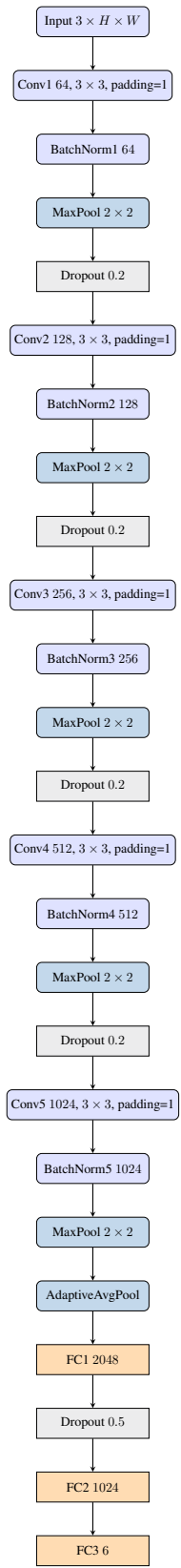
\begin{figure}[ht]
  \centering
  \resizebox{.143\textwidth}{!}{
  \begin{tikzpicture}[node distance=1.7cm]

    \node (input) [layer] {Input $3 \times H \times W$};
    \node (conv1) [layer, below of=input] {Conv1 $64$, $3\times3$, padding=1};
    \node (bn1) [layer, below of=conv1] {BatchNorm1 $64$};
    \node (pool1) [pool, below of=bn1] {MaxPool $2\times2$};
    \node (dropout1) [do, below of=pool1] {Dropout $0.2$};
    \node (conv2) [layer, below of=dropout1] {Conv2 $128$, $3\times3$, padding=1};
    \node (bn2) [layer, below of=conv2] {BatchNorm2 $128$};
    \node (pool2) [pool, below of=bn2] {MaxPool $2\times2$};
    \node (dropout2) [do, below of=pool2] {Dropout $0.2$};
    \node (conv3) [layer, below of=dropout2] {Conv3 $256$, $3\times3$, padding=1};
    \node (bn3) [layer, below of=conv3] {BatchNorm3 $256$};
    \node (pool3) [pool, below of=bn3] {MaxPool $2\times2$};
    \node (dropout3) [do, below of=pool3] {Dropout $0.2$};
    \node (conv4) [layer, below of=dropout3] {Conv4 $512$, $3\times3$, padding=1};
    \node (bn4) [layer, below of=conv4] {BatchNorm4 $512$};
    \node (pool4) [pool, below of=bn4] {MaxPool $2\times2$};
    \node (dropout4) [do, below of=pool4] {Dropout $0.2$};
    \node (conv5) [layer, below of=dropout4] {Conv5 $1024$, $3\times3$, padding=1};
    \node (bn5) [layer, below of=conv5] {BatchNorm5 $1024$};
    \node (pool5) [pool, below of=bn5] {MaxPool $2\times2$};
    \node (adaptivePool) [pool, below of=pool5] {AdaptiveAvgPool};
    \node (fc1) [fc, below of=adaptivePool] {FC1 $2048$};
    \node (dropout6) [do, below of=fc1] {Dropout $0.5$};
    \node (fc2) [fc, below of=dropout6] {FC2 $1024$};
    \node (fc3) [fc, below of=fc2] {FC3 $6$};

    \draw [arrow] (input) -- (conv1);
    \draw [arrow] (conv1) -- (bn1);
    \draw [arrow] (bn1) -- (pool1);
    \draw [arrow] (pool1) -- (dropout1);
    \draw [arrow] (dropout1) -- (conv2);
    \draw [arrow] (conv2) -- (bn2);
    \draw [arrow] (bn2) -- (pool2);
    \draw [arrow] (pool2) -- (dropout2);
    \draw [arrow] (dropout2) -- (conv3);
    \draw [arrow] (conv3) -- (bn3);
    \draw [arrow] (bn3) -- (pool3);
    \draw [arrow] (pool3) -- (dropout3);
    \draw [arrow] (dropout3) -- (conv4);
    \draw [arrow] (conv4) -- (bn4);
    \draw [arrow] (bn4) -- (pool4);
    \draw [arrow] (pool4) -- (dropout4);
    \draw [arrow] (dropout4) -- (conv5);
    \draw [arrow] (conv5) -- (bn5);
    \draw [arrow] (bn5) -- (pool5);
    \draw [arrow] (pool5) -- (adaptivePool);
    \draw [arrow] (adaptivePool) -- (fc1);
    \draw [arrow] (fc1) -- (dropout6);
    \draw [arrow] (dropout6) -- (fc2);
    \draw [arrow] (fc2) -- (fc3);

  \end{tikzpicture}
  }
  \caption{Overview of our detailed model architecture.} 
  \label{fig:modeldetail}
\end{figure}

\tikzstyle{line} = [draw, -latex', line width=0.7pt]